\pdfoutput=1

\documentclass[11pt]{article}

\usepackage{acl}

\usepackage{times}
\usepackage{latexsym}

\usepackage[T1]{fontenc}

\usepackage{graphicx}
\usepackage{microtype}

\usepackage{inconsolata}
\usepackage{subcaption}
\usepackage{float}
\usepackage{natbib}
\usepackage{multirow}
\usepackage{booktabs}

\title{Can LLMs Generate Human-Like Wayfinding Instructions? \\ Towards Platform-Agnostic Embodied Instruction Synthesis}

\author{Vishnu Sashank Dorbala \and Sanjoy Chowdhury \and Dinesh Manocha \\
        University of Maryland, College Park}

\begin{document}
\maketitle
\begin{abstract}
We present a novel approach to automatically synthesize \textit{``wayfinding instructions"} for an embodied robot agent. In contrast to prior approaches that are heavily reliant on human-annotated datasets designed exclusively for specific simulation platforms, our algorithm uses \textit{in-context learning} to condition an LLM to generate instructions using just a few references. Using an LLM-based Visual Question Answering strategy, we gather detailed information about the environment which is used by the LLM for instruction synthesis. We implement our approach on multiple simulation platforms including Matterport3D, AI Habitat and ThreeDWorld, thereby demonstrating its platform-agnostic nature.
We subjectively evaluate our approach via a user study and observe that $83.3\%$ of users find the synthesized instructions accurately capture the details of the environment and show characteristics similar to those of human-generated instructions. Further, we conduct zero-shot navigation with multiple approaches on the REVERIE dataset using the generated instructions, and observe very close correlation with the baseline on standard success metrics ($< 1\%$ change in SR), quantifying the viability of generated instructions in replacing human-annotated data. We finally discuss the applicability of our approach in enabling a \textit{generalizable} evaluation of embodied navigation policies. To the best of our knowledge, ours is the first LLM-driven approach capable of generating \textit{``human-like"} instructions in a platform-agnostic manner, without training.

\end{abstract}
\section{Introduction}

\begin{figure}[t!]
    \centering
    \includegraphics[width=\linewidth]{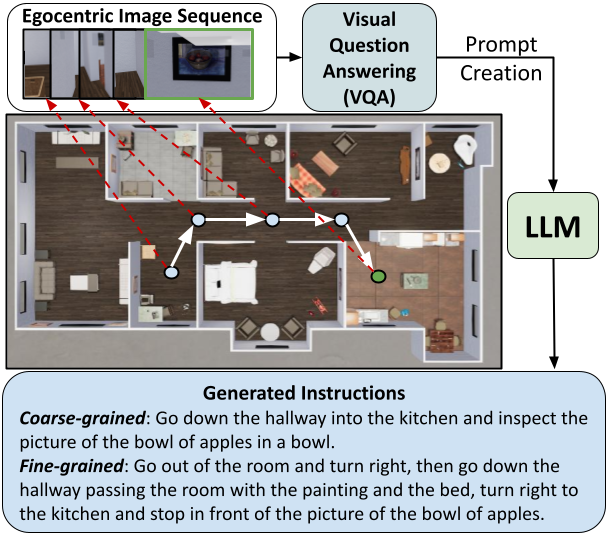}
    \caption{\textbf{Overview}: We use \textit{in-context learning} with an LLM to generate multiple styles of \textit{wayfinding instructions} for embodied navigation. Given \textbf{any} environment, we first gather a set of egocentric images along a path (white arrows), and obtain spatial knowledge via Visual Question Answering. We then condition an LLM on different styles of instructional language (coarse as well as fine grained) via {reference texts}. The figure highlights wayfinding instructions for this environment generated without training on any datasets.}
    \label{fig:cover}
    \vspace{-0.8cm}
\end{figure}

In embodied navigation tasks, language is primarily used to convey \textit{wayfinding instructions} to an agent operating in a simulation platform. These instructions convey the path that the agent should take to reach a target location.
Generating these instructions usually takes place in the form of creating datasets that require several human annotation hours~\cite{REVERIE, R2R, Teach}. 
In addition, the current datasets are exclusive to the embodied simulation platform in which the agent operates, preventing the transfer of instruction-following approaches across platforms. For instance, an embodied agent trained to follow instructions present in the R2R \cite{R2R} or REVERIE \cite{REVERIE} datasets is limited to scenarios (object arrangements and scene layouts) in the Matterport3D ~\cite{matterport, hm3d_habitat} environment, the most commonly used platform for indoor datasets~\cite{jesse_survey}. The scenarios themselves are also limited (around $90$ real-world scans). If its performance needs to be evaluated on another simulation environment such as TDW \cite{TDW} or ProcTHOR \cite{procthor}, the corresponding REVERIE or R2R-style instructions simply do not exist, posing a major hurdle for researchers conducting generalizability experiments to assess the adaptability of their navigation models.
As such, to alleviate these issues, it is important to design an approach for synthesizing wayfinding instructions that are platform-agnostic, and is not cumbersome to generate.


Some recent works have looked at synthesizing instructions from input visual landmarks \cite{synth1, synth2, synth3}. These approaches however are not easily generalizable and require training a separate model for each instruction dataset to infer synthetic instructions. Moreover, they only focus on the Matterport3D environment, as indoor instruction datasets are scarce on other platforms. 

\noindent \textbf{Main Results:} We present a novel approach to synthesize wayfinding instructions for an embodied robot agent.
Figure \ref{fig:cover} presents an overview of our approach. Given a set of egocentric images captured from a simulator, we perform Visual Question Answering to gather information about the scene, and use this to condition an LLM with reference texts to generate different styles of instructions.
The novel components of our work include:
\begin{itemize}
    \item We present a novel platform-agnostic, non-training based approach to synthesize wayfinding instructions of multiple styles.
    \item We use the \textit{in-context learning} capabilities of LLMs to perform instruction synthesis in a few-shot manner. Our method only requires a few samples of reference wayfinding text to produce human-like instructions in multiple simulation platforms.
    \item We subjectively validate generated instructions across multiple simulation platforms via a user study and infer that $83.3\%$ of users find the instructions accurately capture details of the environment, and exhibit human-like characteristics. 
    \item Finally, we evaluate the effectiveness of our generated instructions on the REVERIE vision-and-language navigation (VLN) task. The performance of three zero-shot VLN approaches, evaluated using standard VLN success metrics, was comparable to established baselines, highlighting the efficacy and practical utility of LLM-generated instructions in navigation tasks.
\end{itemize}
In contrast to prior work which is limited to a single simulation platform and instruction style, we use in-context learning in LLMs to achieve \textit{instruction synthesis} of multiple styles on different embodied simulation platforms, including Matterport3D, AI Habitat and ThreeDWorld. Our evaluation both via a user study and navigation performance indicates that the synthesized instructions are sufficiently representative of human-like texts for them to be used as a scalable alternative for generating instructions for embodied navigation tasks.




\section{Approach}

Our approach consists of two components. First, we perform Visual Question Answering (VQA) on egocentric images taken along an agent's path in a simulation environment. This gives us spatial knowledge about the scene. Next, we combine this spatial knowledge with a few reference \textit{wayfinding instructions} in an in-context learning \cite{in-context1} prompt to condition an LLM for synthesizing instructions that would lead the agent to the target location.

\subsection{Extracting Spatial Knowledge: LLM + BLIP}
\label{sec: spatial}

\begin{figure}[t!]
    \centering
    \includegraphics[width=\linewidth]{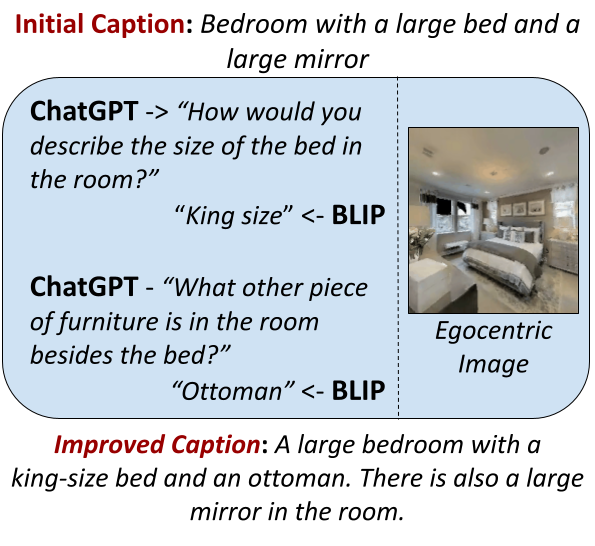}
    \caption{\textbf{Extracting Spatial Knowledge}: We use the GPT-3.5-turbo along with BLIP to maximize knowledge captured from an image, similar to ChatCaptioner \cite{chatgptblip}. We notice that adding more detail to the captions helps improve the quality the final instruction by filtering out unnecessary information. More details about this are in Appendix \ref{app: fsps}.}
    \label{fig:conversation}
    \vspace{-0.5cm}
\end{figure}

Paths in simulated environments describe a navigable route for an embodied agent to get from one point to another. In our approach, given any embodied simulator, we first generate random paths. We then obtain a discrete set of egocentric images $\mathcal{I}$ uniformly sampled on this path.

We then perform VQA on the images in $\mathcal{I}$, to gather information about the environmental artifacts on the path. Following a similar approach presented in ChatCaptioner \cite{chatgptblip}, we maximize the knowledge obtained from each image by gathering insights via a conversation in a Chain of Thought manner \cite{cot} between GPT-3.5 \cite{chatgpt} and BLIP \cite{blip2} (Figure \ref{fig:conversation}). We notice that this gives us more detailed descriptions of each image, improving the quality of the generated instruction.

\subsection{Synthesizing Wayfinding Instructions via In-Context Learning}

\begin{figure}[t!]
    \centering
    \includegraphics[width=\linewidth]{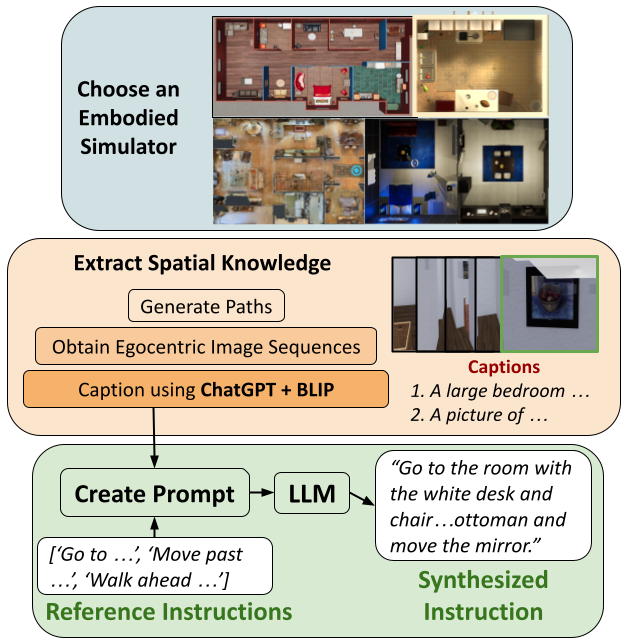}
    \caption{Given any embodied simulator, we synthesize multiple styles of wayfinding instructions for agents. Spatial knowledge is first mined from egocentric images $\mathcal{I}$ captured using the LLM and BLIP. These captions are fed into a prompt along with a few reference examples representing the desired instruction style. Finally, the LLM is conditioned with this prompt to generate a human-like instruction in the style of the reference text, using the captioned information.}
    \label{fig:gpt}
    \vspace{-0.5cm}
\end{figure}

We condition GPT-3.5-turbo-instruct to generate suitable wayfinding instructions for navigation. Figure \ref{fig:gpt} illustrates this approach. Captions obtained for images in $\mathcal{I}$ along with \textit{reference texts} providing context on the desired instruction style are used to create a prompt for the LLM.  We experiment with reference instructions taken from two datasets with contrasting styles; \textbf{R2R} \cite{R2R}, which has more detailed, \textit{fine-grained} human annotations, and \textbf{REVERIE} \cite{REVERIE}, which has instructions that are abstract and \textit{coarse-grained}.

We also observe that adding more information about the instruction style itself helps further finetune the outcome. For instance, in the REVERIE dataset \cite{REVERIE}, almost all instructions end by describing a task with the target object (\textit{`turn the faucet'} for example). Adding this information as an additional constraint helps further finetune the LLM output. More details about this are provided in appendix \ref{app: fsps}.




\section{Evaluation \& Results}
\begin{table*}[t!]
  \centering
  \setlength{\tabcolsep}{6pt} 
  \begin{tabular}{@{}cccccccccc@{}} 
    \hline
    \multirow{3}{*}{Approach} & \multicolumn{3}{c}{Original} & \multicolumn{3}{c}{Generated (Central)} & \multicolumn{3}{c}{Generated (Panoramic)} \\
    \cmidrule(lr){2-4}\cmidrule(lr){5-7}\cmidrule(lr){8-10}
    & SR $\uparrow$ & OSR $\uparrow$ & SPL $\uparrow$ & SR $\uparrow$ & OSR $\uparrow$ & SPL $\uparrow$ & 
    SR $\uparrow$ & OSR $\uparrow$ & SPL $\uparrow$ \\
    \hline
    Clip-Nav &  6.57 & 28.68 &  0.06 & 5.98 & 26.69 & 0.05 & 5.57 & 26.09 & 0.05 \\
    \hline
    Seq-CLIPNav & 14.92 &  24.46 & 0.15 & 13.94 & 21.51 & 0.14 & 11.35 & 23.10 & 0.13 \\
    \hline
    GLIP-Nav & 16.87 & 32.56 & 0.18 & 16.32 & 33.23 & 0.18 & 14.18 & 29.87 & 0.15 \\
    \hline
  \end{tabular}
  \caption*{\textbf{Results}: We evaluate zero-shot VLN models by replacing REVERIE's human-annotated instructions with instructions generated by our approach. Notice the similar performance on each VLN model across all metrics. There is a noticeable drop in using panoramic frames over central frames, and this could be attributed to condensing copious amounts of scene information into a single sentence (See Appendix \ref{app:frame_select}). We can positively infer from the minimal difference in SR, OSR, and SPL values that our approach can generate instructions that can indeed serve as a good replacement to human-annotated data.}
  \label{tab:results}
  \vspace{-0.3cm}
\end{table*}

In this section, we discuss our evaluation strategy and present results.

\subsection{Qualitative: User Study}
We conduct a user study to evaluate the quality of the generated instructions. Participants are first shown a video of a random path taken from one of 3 different simulators (Matterport3D, AI Habitat, ThreeDWorld). Using an instruction of either a REVERIE or R2R style as reference they are asked to come up with a stylistically similar instruction for the video. We then show them the generated instruction, and ask them a few questions about correlation. 

We infer that $83.3\%$ of users believe that the generated instruction captured details of the environment to more than a decent level of accuracy, and that a majority of $73.3\%$ believed that the agent could reach the target room by following the generated instruction. Further, $16.5\%$ of participants reported seeing ghost objects, indicating generation is sensitive to the captioning scheme. Conversely, $43.3\%$ of participants believed the instructions were different from what they wrote. This indicates that the vocabulary people use to describe a path may significantly vary from the vocabulary used in the generated instruction. This however is not an indicator of instruction quality, as the difference is in alternate landmarks being used guide the agent along the same path. This is further highlighted in the navigation results presented below.
More details are in Appendix \ref{app:user_study}.

\subsection{Quantitative: Embodied Navigation}
\label{sec:results}
Our evaluation setup is simple. We first implement a zero-shot navigation scheme using the original instructions provided in REVERIE, a popular VLN dataset. We then replace the original instructions with instructions generated by our approach, and run the navigation scheme again. A similar performance would indicate that the generated instructions can indeed serve as a replacement to human-annotated data.

REVERIE is based on the Matterport3D simulator, which contains real-world captures of household environments. We look at 3 zero-shot VLN approaches - 1) \textbf{CLIP-Nav} \cite{clipnav}, which uses CLIP \cite{CLIP} to ground target instructions to a scene to drive the agent's navigation policy, 2) \textbf{Seq-CLIP-Nav}, an extension of this approach that also performs backtracking (see Appendix \ref{app:embodied_nav}), and 3) \textbf{GLIP-Nav}, which we introduce as a GLIP \cite{GLIP} based variant of Seq-CLIP-Nav. More details about these approaches are in Appendix \ref{app:embodied_nav}. 

As Matterport3D provides panoramic images, we consider two possibilities for extracting spatial knowledge (see Appendix \ref{app:frame_select}); The \textbf{Central Caption}, where only the images in the direction of the agent's heading are captioned, and the \textbf{Panoramic Caption}, where the entire panorama ($4$ images) is captioned and summarized to obtain an instruction.

\noindent\textbf{Experiment Details}: We employ $3$ standard VLN evaluation metrics \cite{VLN-eval} to measure performance across each navigation approach - 1) \textbf{SR}, which is the \textbf{S}uccess \textbf{R}ate determining when the agent has successfully reached the target location; 2) \textbf{OSR}, the \textbf{O}racle \textbf{S}uccess \textbf{R}ate, for when the agent successfully reached the target location once, but overshot and stopped elsewhere, and 3) \textbf{SPL}, which measures efficiency of \textbf{S}uccess weighted by \textbf{P}ath \textbf{L}ength.
The results table compares the performance of the generated instructions with the original ones on the zero-shot VLN approaches. 

\noindent We make the following key \textbf{inferences} - 

\noindent\textbf{Automated Instruction Generation}: A key observation is that embodied agents equipped with LLM-generated instructions perform almost equally well compared to when they are provided with human annotated instruction. This has practical implications for researchers working on embodied navigation, where such instruction data is limited and hard to annotate. Creating large-scale instruction datasets is challenging, often needing simulator-specific annotation tools, which cannot be easily transferred. To this end, our study presents a good alternative in leveraging off-the-shelf LLMs as a wayfinding instruction generation tool. \\
\noindent\textbf{Central vs. Panoramic Captions in MP3D}: We observe that the performance of the central caption approach is generally higher than that of the panoramic caption approach. We believe this to be due to instruction quality being affected by two reasons \emph{---} 1) Captioning each image of the panorama and summarizing it leads to excess information at each step and 2) The central caption approach implicitly contains the information in the heading of the target, leading to more direct instructions. \\
\noindent\textbf{Cross-Platform Scalability}: Our approach is platform-agnostic, and can be applied to generate instructions across embodied simulation platforms, whether they are discrete, continuous, photorealistic, or not. The user study validates this, where users across simulator types believed that the generated instructions captured details of the environment and could lead the agent to the target location. We believe that the embodied navigation community can significantly benefit from this, enabling researchers to conduct cross-platform generalizability experiments without relying on the availability of platform-specific human-annotated data. \\
\noindent\textbf{Improved Instruction Quality}: We notice that human-annotated instructions in REVERIE sometimes tend to be unnatural and lacking in terms of sentence construction. As these annotations are crowdsourced, this can be attributed to human error. It is often in these cases that the embodied agent fails to reach it's target location, due to poor annotation leading to inferior grounding scores. LLM-generated instructions on the other hand are almost always well structured, containing specific objects and waypoints leading up to a target location; a direct consequence of our prompting strategy. Some of these cases are discussed in appendix \ref{app:instr_qual}.


\section{Discussion: Evaluating Generalizability of Embodied Navigation Policies}

The overarching motive of our work is to construct a \textbf{generalist navigation agent} that performs \textit{consistently} irrespective of the environment that it is present in. Current approaches to solve this task are limited to evaluation on human-annotated datasets created specifically for a particular simulator, be it MP3D \cite{matterport}, AI Habitat \cite{hm3d_habitat}, RoboThor \cite{ai2thor} etc.. While some methods claim generalizability \cite{parksurvey}, they back their claims by showing improved performance on unseen subsets of a dataset on the \textit{same} simulator, rather than measuring performance \textit{across} simulators. For a true measure of generalizability, we believe it is necessary to measure the navigation performance of agents that aren't bounded to a particular dataset.

In this direction, our approach solves a crucial data procurement problem in providing a simple method to generate human-like instructions across simulation platforms. In doing so, we \textit{empower} resource-constrained researchers to create their own datasets for generalizable experiments on their navigation models; therein presenting the true novelty of our work.

Current datasets cover a wide range of language-guided navigation scenarios, ranging from initial-instruction based guidance (fine and coarse-grained) to oracle and dialogue based navigation that provide verbal human assistance \cite{jesse_survey}. There also exist several outdoor datasets including Touchdown \cite{touchdown}, Talk2Nav \cite{talk2nav} and StreetNav \cite{streetnav}, where the beyond the instruction, the structure and semantics of the scene are drastically different from indoors. To account for the diversity and measure true generalizability, we propose integrating our scheme for synthesis to measure the \textbf{\textit{robustness}} of navigation policies in two ways as follows:- 

\begin{itemize}
    \item \textbf{Cross-Platform Generalizability}: In the first experiment, we gather a set of instruction-path pairs across simulators to train a cross-platform model for a generalist navigation agent. Consistent performance on each simulator present in the dataset during inference would indicate that the navigation policy is \textit{\textbf{globally}} robust with low bias towards a specific simulator.
    \item \textbf{Intra-Platform Generalizability}: In the second experiment, we measure the agent's performance \textit{within} different generated datasets on the same simulator. Unlike data augmentation approaches in the past \cite{li2022envedit} that seek to improve the agent's performance with generated instruction-path data, our objective is measure \textit{consistency} in performance across multiple instruction-path ``datasets'' generated in the same environment. This consistency would indicate that the navigation policy is \textit{\textbf{locally}} robust, with low bias towards a specific type of scene or region within the simulator.

\end{itemize}

A generalist navigation agent would have a policy that is both globally and locally robust. Our approach paves the way to measure this robustness for a fair evaluation of state-of-the-art embodied navigation policies. 

\section{Conclusion}
We present a simple, cross-platform approach to synthesize multiple styles of wayfinding instructions for embodied navigation. Our approach requires no training and instead utilizes an LLM with in-context learning to produce instructions across multiple simulation platforms. We verify the quality of the instructions generated both via a user study and by evaluating zero-shot VLN performance. From these evaluations, we positively infer that our LLM-generated instructions are a good replacement to human-annotated ones, and further, that our approach provides for a scalable and accessible solution for creating wayfinding instructions. We finally touch upon how our approach can be used for measuring the key quality of \textit{robustness} while evaluating language-guided navigation policies; a defining metric to evaluate a generalist navigation agent.

\section{Limitations and Future Work}
While our approach is platform-agnostic, the quality of the generated instructions is very sensitive to the individual modules that drive our scheme. Poor spatial knowledge extracted from performing VQA would directly affect the quality of the caption. In some preliminary experiments, we notice this behavior on some images taken from the VirtualHome \cite{virtualhome} embodied simulator, which has non-photorealistic environments. Using LLaVA \cite{llava} for VQA seems to create ghost objects and artifacts when asked to describe a scene leading to poor instructions. In contrast, it performs well with real world images taken from Matterport3D. We believe this poor performance might be because large captioning models such as LLaVA are trained on an abundance of real world data, and may contain fewer if not any simulation or non-photorealistic images. Secondly, during the synthesis stage, we present the LLM with examples from the instruction style that we wish to obtain. The generated instructions can sometimes contain the direct words or language used in these reference examples. As such, we believe it is necessary to explicitly specify in the prompt that the LLM uses only the captions and not the reference texts for generation.
In the future, we intend to use our approach to implement a \textit{generalist navigation agent} and study its performance in terms of \textit{consistency} across various embodied simulation platforms.

\section{Ethics Statement}
Equipping embodied agent with LLM-generated instructions to perform navigational tasks is a step towards cohesive human-robot collaboration. While the end goal is to make such systems fault-tolerant and error-free, we may not want an agent to perform certain actions that it is unsure of. However, currently there seems to be a gap in the language interpretation capabilities of the agent especially in complex scenarios. \\      
Our user study protocol was approved by Institutional Review Board and we do not collect, share or store any personal information of the participants.

\section{Acknowledgments}
This work was supported in part by ARO Grants W911NF2110026, W911NF2310046,  W911NF2310352  and Army Cooperative Agreement W911NF2120076. We would also like thank Niall L. Williams for his creative insights.
\bibliography{custom}

\clearpage
\appendix

\begin{figure*}[t!]
\centering
     \includegraphics[width=1.0\textwidth]{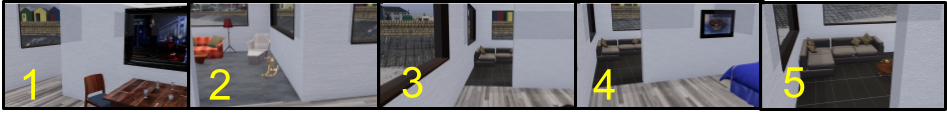}
      \caption{Egocentric Image Sequence from a path in ThreeDWorld \cite{TDW}}
       \label{fig:tdw_sequence}
\end{figure*}

\begin{figure*}[t!]
\centering
     \includegraphics[width=1.0\textwidth, height=2cm]{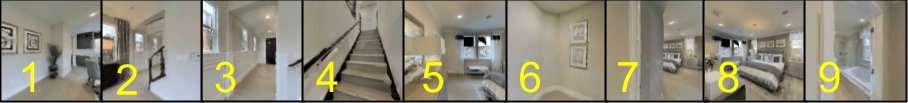}
      \caption{Egocentric Image Sequence from a path in AI Habitat \cite{hm3d_habitat}}
       \label{fig:habitat_sequence}
\end{figure*}

\begin{figure*}[t!]
\centering
     \includegraphics[width=1.0\textwidth, height=2cm]{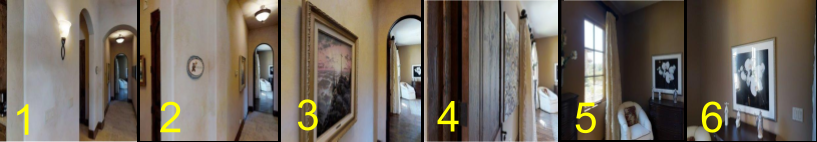}
      \caption{Egocentric Image Sequence from a path in Matterport3D \cite{matterport}}
       \label{fig:matterport_sequence}
\end{figure*}

\section{In-Context Learning Strategies} 
\label{app: fsps}

In this section, we discuss some strategies we employ to get the best possible wayfinding instruction. A prompt template is presented to the LLM as -
\begin{quote}
   \textit{ "A robot agent at home sees a sequence of egocentric images with the following frame descriptions. \\
    Frame 0: <\textit{Caption 1}> \\
    Frame 1: <\textit{Caption 2}> \\
    ... \\
    Frame n: <\textit{Caption n}> \\
    Reference Texts: ['Go to ...', 'Move past ...', Walk ahead ... ] \\
    Write an concise instruction \textit{in the style} of the Reference Texts that would get the robot from Frame 0 to Frame \textit{n}. \\}
    <\textit{Additional constraints on the instruction style}>"
    \label{prompt}
\end{quote}

Each caption in this template is obtained using the LLM + BLIP strategy outlined in section \ref{sec: spatial}.

\subsection{Influence of LLM + BLIP}
\label{app:llm_blip}
An example egocentric image sequence of a path taken in the TDW simulator \cite{TDW} is shown in figure \ref{fig:tdw_sequence}. Using the LLM + BLIP approach discussed in section \ref{sec: spatial}, we get the following captions for each image in \ref{fig:tdw_sequence}.

\begin{enumerate}
    \item \textit{The image depicts a computer screen showing a colorful video of a man that is being displayed on a television. There is also a chair visible in the image besides the television.}
    \item \textit{The image contains a small chair made of fabric, in colors of red, white and gray. There is another object present in the image, but it is not clear what it is.}
    \item \textit{The image is of a living room with brown furniture and no decorations on the walls. There are no people present in the living room.}
    \item \textit{The image depicts a room with a gray couch located against a wall. There is a small television mounted on the wall.}
    \item \textit{The image features a computer screen displaying a website, with a couch visible in the background. A plant is placed on a table next to the computer. No other objects are visible on the table.}
\end{enumerate}

Passing these Frame descriptions to the prompt in \ref{prompt} gives us a REVERIE-like instruction as,

\begin{quote}
    \textit{Go to the living room, then move to the room with the gray couch and turn off the television mounted on the wall.}
\end{quote}

and R2R-like instruction as,

\begin{quote}
    \textit{Go from the computer screen to the chair, then past the object in the background and into the living room. Walk past the blue furniture and turn right towards the gray couch. Finally, stop in front of the table with the plant and view the website on the computer screen.}
\end{quote}

Notice the level of detail added by \textit{insights} that LLM gathers by conversing with BLIP. This inturn gives more information for GPT-3.5-Turbo-Instruct to use for generating a final instruction from the frames.

\begin{figure}[H]
    \centering
    \includegraphics[width=\linewidth]{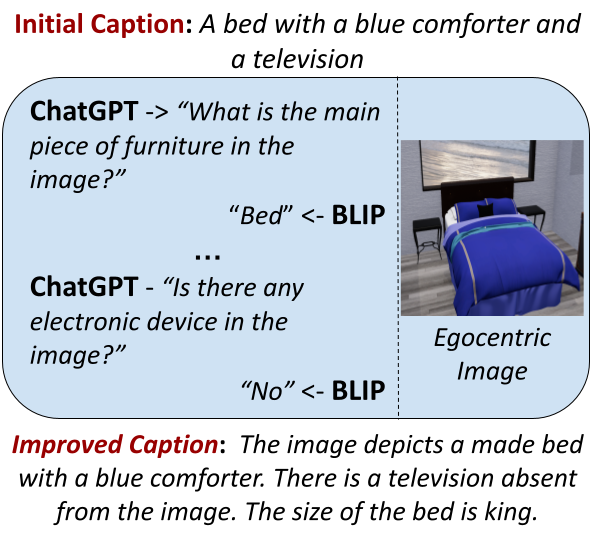}
    \caption{\textbf{LLM + BLIP}: Notice the initial BLIP caption mentions a television in the image, even when it is not present. When the LLM asks for the presence of electronic items in the room, BLIP answers no, which leads to the refined caption preventing misinformation.}
    \label{fig:blipchat-refine}
\end{figure}

We also experiment with using only BLIP captions with object and room queries, without the LLM. The REVERIE-like instruction in this case looks like, 

\begin{quote}
\textit{Go to the living room on level 0 and turn off the television by the couch and the table.}
\end{quote}

and the R2R-like instruction is,

\begin{quote}
\textit{Start in the kitchen and go up the stairs on the left. Turn right at the top of the stairs and then go past the round table and chairs and stairs. Keep walking until you see the two small tables on the rug and then turn left. Go down the hallway keeping the wall on your left and stop in front of the door on your right with the treadmill. Turn left and you will see the living room with a computer screen containing a picture of a couch and a table.}
\end{quote}

While the REVERIE-like instruction is still usable, notice the R2R-like instruction tends to be nonsensical with ghost objects such as \textit{stairs} and \textit{treadmill} in the caption. It also contains bad directions.
We observe this phenomenon in multiple cases, and Figure \ref{fig:blipchat-refine} showcases how the conversation with the LLM improves the initial captions to remove ghost objects and prevent misinformation.

Thus, we infer that using an LLM with BLIP to provide more detail about the environment is important when it comes to finally generating more meaningful instructions.



\subsection{Empirical Information on Instruction Styles}

We utilize factual knowledge about R2R and REVERIE instruction styles to finetune the LLM prompt.

\subsubsection{Additional Constraints for R2R}

Upon inspection, we observe that R2R instructions are usually 2 or more sentences long, attributed to longer path lengths. Further, in the R2R paper, the authors mention that they ask annotators to ``write directions so that a smart robot can find the goal location after starting from the same start location", and are told that it is not necessary to follow the path, but only to reach the goal. We incorporate this information to append our prompt:-
\begin{quote}
    ``Write directions so a smart robot can find the final frame after starting from the same starting frame. You do not have to use information in the frames, and just need to reach the goal location."
\end{quote}

\subsubsection{Additional Constraints for REVERIE}

REVERIE instructions are concise, and talk only about the goal location.
Clip-Nav \cite{clipnav} studies REVERIE in detail and empirically deduces that most instructions can be broken down into \textit{navigation} and \textit{activity} components, with the conjunction \textit{and} between them. We utilize this information to add the following to our prompt:-

\begin{quote}
    "The instruction must be a single sentence long, ending with a task related to an object in the final frame, and must be less than 20 words."
\end{quote}

\section{Evaluation Details}

\subsection{Simulator Implementations}

We implement our approach on 3 different simulation platforms, namely AI Habitat \cite{hm3d_habitat}, Matterport3D \cite{matterport} and ThreeDWorld (TDW) \cite{TDW}. Egocentric image sequences for these simulators are presented in Figure \ref{fig:tdw_sequence}, Figure \ref{fig:habitat_sequence} and Figure \ref{fig:matterport_sequence} respectively.
Depending on the type of simulator, we revise our strategy for extracting sequences as listed below - 
\begin{itemize}
    \item Environments in the \textbf{Matterport3D} simulator are taken from real world scenes and provide fully connected graphs whose nodes represent 360 panoramas. Given two nodes from the connected graph, we compute a path between them as a sequence of nodes. To compute captions, we either consider the central frame or the entire panorama (described in Appendix \ref{app:frame_select}). The path contains discrete ``hops" of in the form of images, which gives us our image sequence.
    \item \textbf{AI Habitat} has continuous 3D reconstructions of real world household environments. To obtain a path, we first sample two navigable points in the environment and compute the shortest distance between them. Then, to obtain a discrete sequence of images, we sample images at a uniform interval along the path.
    \item \textbf{TDW} is a photorealistic simulator that is capable of procedurally generating new environments. We make use of this simulator to test the robustness of our approach in non-real world environments. We obtain our image sequence in the same manner as AI Habitat.
\end{itemize}

For the user study, we sample 100 paths of varying lengths from each of these simulators, randomly choosing from environments they offer. We then use our approach on these paths to generate instructions in a platform-agnostic manner.


\subsection{Qualitative Analysis - User Study Details}
\label{app:user_study}

Each user is presented with a random image sequence chosen from a bank of sequences gathered from the 3 different environments. This allows for us to evaluate the generated instruction across multiple platforms. We observe a consistent performance across simulators, leading us to establish the platform-agnostic nature of our instruction synthesizer.

Our study was aimed at quantifying the usability of generated instructions in guiding an embodied agent in the environment. In this direction, we first presented the user with video of an egocentric image sequence chosen from a random simulation platform. After being shown examples of fine or coarse grained instructions, the users were asked to provide an instruction describing the robot's path in that style. Finally, the participant is shown the synthesized instruction for the same sequence and is asked comparative questions highlighted in figure below.

\begin{figure}[h!]
    \centering
    \includegraphics[width=\linewidth]{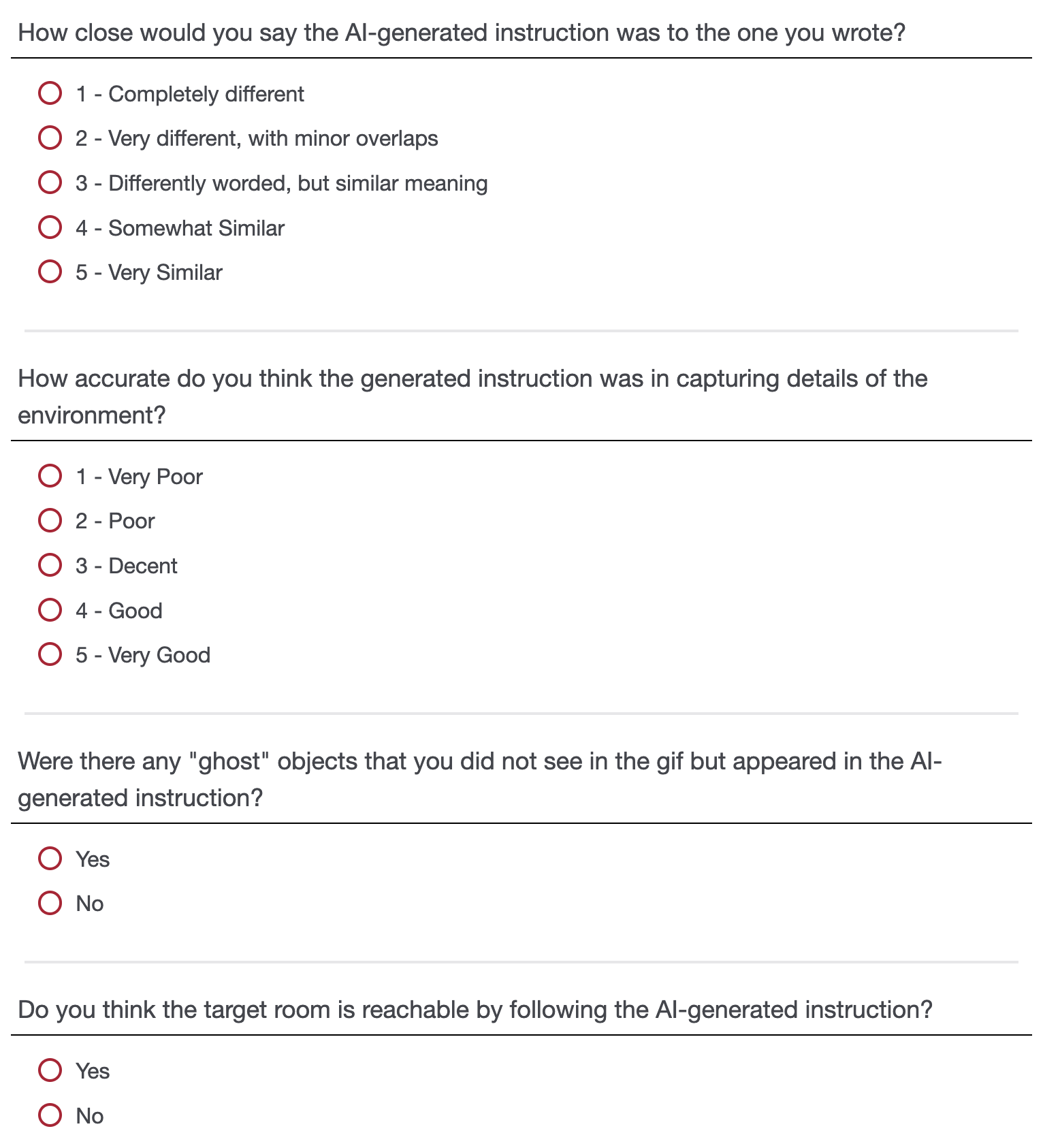}
    \caption*{\textbf{Our User Study}. The participant is asked questions on the quality of the generated instructions and about how much it compares with the instruction that they wrote.}
    \label{fig:survey}
    \vspace{-0.3cm}
\end{figure}

Each question aims to tackle a different comparative perspective. The first question seeks to find out if the generated instructions are similar to what the user has written down.
The second question asks if the generated instructions accurately capture details of the environment. The third queries about the robustness of generation by asking if the participant has noticed any ghost objects or artifacts. Finally, we ask if the user thinks an embodied agent could reach the target location by following the generated instruction. 

Out of a total of 30 participants, $83.3\%$ believed the instruction captured details of the environment to a more than decent level of accuracy. A majority ($73.3\%$) of these users also believed that the agent could reach the target room by following the generated instruction. A lower percentage of participants ($16.5\%$) reported seeing ghost objects, which indicates either that some people may have missed objects in the video, or that the generated instruction is sensitive to the captioning scheme.

Conversely, $43.3\%$ of participants believed that the instructions generated were either very different from what they wrote, or had minor overlaps. We can infer from this that the vocabulary people use to describe a path may significantly vary from the generated instruction. However, this does not necessarily mean that the agent would not be able to follow the generated instruction to reach the target location, as it would use alternate references or landmarks to get there.

Our study was determined exempt by our institution's IRB. All of the participants voluntarily chose to participate in it.

\subsection{Quantitative Study - Zero-Shot Embodied Navigation}
\label{app:embodied_nav}


\subsubsection{Dataset and Navigation Setup Details}
We run navigation experiments on the REVERIE dataset, which tackles vision-and-language navigation (VLN) using coarse-grained instructions. Instructions in REVERIE have been human-annotated, where the annotator is asked to write a high-level instruction describing how to get to the target location after being shown a path in the Matterport3D environment. Each path is discrete, i.e., it consists of a set of panoramic images or nodes along which the agent ``\textit{hops}". The nodes inturn consist of $4$ views covering a $360$ degree view of the agent.

We consider a generalizable, zero-shot case, where the agent is dropped in an environment that it has no knowledge of, and is given an instruction that it must follow to get to a target location. This setting is in line with our ultimate goal of developing a generalist embodied navigation agent, which is able to function without any supervision in an unseen environment. We opt to use the unseen validation split of the REVERIE dataset for evaluation, which contains environments that the agent would not see in the training split. It contains $504$ paths, which was deemed sufficient for showcasing zero-shot navigation prowess using the generated instructions.

\textbf{CLIP-Nav} \cite{clipnav} uses CLIP to make grounding decisions for navigation. The instruction is first broken down into a Navigation Component (NC) and an Activity Component (AC). The NC contains information about getting to the target location, while the AC containing the activity that the agent is expected to perform is disregarded. The NC is further broken down into noun phrases using GPT-3.5-turbo, which are then grounded using CLIP with each of the 4 images captured by the agent from its panoramic view. The agent takes the direction of the highest CLIP grounding score. \\
\textbf{Seq-CLIP-Nav} extends this to incorporate backtracking. Backtracking refers to when the agent falls back or ``backtracks" a few nodes when it determines that it has taken the wrong path. 

We also ablate with \textbf{GLIP-Nav}, a variant of Seq-CLIP-Nav we introduce, where CLIP is replaced with GLIP \cite{GLIP} for obtaining grounding scores.

\subsubsection{Matterport3D: Frame Selection}
\label{app:frame_select}
REVERIE provides a set of panoramic images taken from Matterport3D that forms a path corresponding to each instruction. The annotator is provided with this whole panoramic view at each step. To incorporate our generation approach here, we consider two variations.

\textbf{Central Caption}: We hypothesize that the central frame contains the most immediate and critical information required for the embodied agent to perform its next set of actions. To this end, we caption only the central frames (i.e., the image in the direction of the agent's heading) of the entire path sequence to generate the instruction.
 
\textbf{Panoramic Caption}: Here we caption each image of the entire panorama (4 frames), and summarize the individual captions using the LLM. We perform this over the entire path sequence to generate the instruction. Although the panoramic sequence contains more semantic information over the single (central) frame, note that each instruction is only a single sentence, and compressing all the information of a scene (be it the target or an image along the path) is non-trivial, if the instruction has to be of a suitable length. 

During the panoramic-frame case, we use the LLM to summarize the set of captions obtained $4$ $90$ degree views around the agent. Each caption in this set is obtained using the LLM + BLIP approach discussed in section \ref{sec: spatial}. The prompt for this is -  

\begin{quote}
   \textit{ "I see a panoramic view with the following descriptions. \\
    North: <\textit{Caption 1}> \\
    East: <\textit{Caption 2}> \\
    South: <\textit{Caption 3}> \\
    West: <\textit{Caption 4}> \\
    Summarize these descriptions into a single description using less than $20$ words.}"
    \label{summary_prompt}
\end{quote}

\subsubsection{Inferences on Generated Instructions}
\label{app:instr_qual}

In addition to the results presented in section \ref{sec:results}, we also measure the \textit{average pairwise cosine similarity} using MiniLM-V6 \cite{sentence-bert} between the human-annotated instructions and the generated instructions.

For the central-caption case, we get a score of $0.476$, and for the panoramic-caption case, we get $0.433$, on a scale of $-1$ to $1$. From the overall positive correlation, we can infer that the generated instructions tend to be similar to the human-annotated ones on average. Some individual cases of extreme difference are discussed below.

In a low cosine similarity example, consider
\begin{quote}
    \noindent\textbf{Human-Annotated}: \textit{"Walk to the bottom of the stairs leading to the level 1 hallway and find the bottommost stair"}\\
    \noindent\textbf{Generated}: \textit{"Move from bedroom to kitchen, turn off faucet."} \\
    \noindent\textbf{Similarity}: $0.0850$
\end{quote} 

Notice that the human-annotated instruction presents a unique situation to the agent where it is expected to find the \textit{bottommost stair}. In contrast, the generated instruction asks the agent to move to the kitchen, which is near the vicinity of the staircase in this environment.
While the cosine similarity might be low, a generalist agent would still be able to reach the target location with the given instruction since it references other elements (``the faucet" here) in the scene. Note that VLN tasks deal with the agent reaching a target location, and not with what it needs to do once it gets there.

In a high cosine-similarity example, consider,
\begin{quote}
    \noindent\textbf{Human-Annotated}: \textit{"Go through the nearest bedroom to the bathroom on the first floor and turn on the faucet on the rightmost"}\\
    \noindent\textbf{Generated}: \textit{"Go to the bedroom and turn off faucet."} \\
    \noindent\textbf{Similarity}: $0.820$
\end{quote} 

Observe that a high cosine similarity does not necessarily mean that the generated instruction is of good quality. In this example, notice that the human annotator asks the agent to enter the bathroom after going through the bedroom to turn off the faucet. The generated instruction however entirely misses out on entering the bathroom, which would cause an agent to incorrectly look for a faucet in the bedroom.

These are however one-off cases; we observe that most generated instructions tend to closely follow or paraphrase human-annotations.
For instance, consider, 

\begin{quote}
    \noindent\textbf{Human-Annotated}: \textit{"Go to the bathroom on level 1 and wipe off the faucet"}\\
    \noindent\textbf{Generated}: \textit{"Go to the wooden room on level 1, turn off faucet in the bathroom."} \\
    \noindent\textbf{Similarity}: $0.885$
\end{quote} 

Both these instructions ask the agent to go to the bathroom on level 1 to execute a task.

\section{Related Work}

\subsection{Embodied Instruction Synthesis}
Embodied or Vision-and-Language Navigation deals with the problem of navigating an agent in unseen photorealistic environments and adhering to language instructions. These wayfinding instructions are usually human annotated as part of datasets \cite{ku2020room, qi2020reverie, anderson2018vision, krantz2020beyond}, and can roughly be categorized into coarse and fine-grained \cite{jesse_survey} based on their level of detail.
As these datasets are exclusive to the environments that they are created in, generalizing them to other new or procedurally generated environments presents a unique challenge. 
Most prior work on instructions synthesis \cite{li2022envedit} has mostly been tailored toward data augmentation. \cite{instrgen1} presents a counterfactual reasoning approach to generate instructions, but ultimately requires the model to be trained on the R2R \cite{R2R} dataset. \cite{synth1, marky1} present imitation learning models that are trained on datasets, and use the augmented instructions to improve navigation performance. More recently \citet{lana} presents a navigation agent which is able to not only execute human-written navigation commands, but also provide route descriptions to humans. These approaches are limited to a few datasets and have cumbersome training procedures. In contrast, our approach can generalize over multiple styles of instructions, over multiple simulation platforms without requiring a dataset.

\subsection{LLMs for Embodied Robot Navigation}
Vision-and-Language Navigation (VLN) has been a popular task in Embodied AI, with several pre-LLM era approaches using BERT features, such as VLN-BERT \cite{vln-bert, vln-trans}, VilBERT \cite{vilbert}, and Airbert \cite{airbert}.  
Recent work has used LLMs being for this task \cite{vlmaps, navgpt}, especially in a zero-shot setting \cite{l3mvn, clipnav}. While \cite{lmnav} leverage GPT-3.5 \cite{brown2020language} to identify landmarks, \cite{zhou2023esc} and \cite{LGX} use an LLM for commonsense reasoning between objects and targets to facilitate navigation. With LLMs being increasingly used in several embodied AI frameworks beyond navigation \cite{embodiedgpt, inner}, utilizing them for instruction generation allows for easier integration and testing at a system level. Finally, March-in-Chat (MiC) \cite{march} can talk to the LLM on the fly and plan the navigation trajectory dynamically. 

\end{document}